\pdfoutput=1

\documentclass[sigconf]{aamas}  

\usepackage{booktabs}

\usepackage{verbatim}
\usepackage{graphicx}
\usepackage{amsmath}
\usepackage{amsfonts}
\usepackage{amssymb}
\usepackage{algorithm}
\usepackage{algpseudocode}
\usepackage{pbox}
\usepackage{epsfig}
\usepackage{xcolor}
\usepackage{soul}
\usepackage{bm}
\usepackage{multirow}
\usepackage{tabularx}

\setcopyright{ifaamas}  
\acmDOI{doi}  
\acmISBN{}  
\acmConference[AAMAS'19]{Proc.\@ of the 18th International Conference on Autonomous Agents and Multiagent Systems (AAMAS 2019), N.~Agmon, M.~E.~Taylor, E.~Elkind, M.~Veloso (eds.)}{May 2019}{Montreal, Canada}  
\acmYear{2019}  
\copyrightyear{2019}  
\acmPrice{}  

\begin{document}

\title{Learning Curriculum Policies for Reinforcement Learning}  




%
\author{Sanmit Narvekar}
\affiliation{%
  \institution{Department of Computer Science \\ University of Texas at Austin}
}
\email{sanmit@cs.utexas.edu}
\author{Peter Stone}
\affiliation{%
  \institution{Department of Computer Science \\ University of Texas at Austin}
}
\email{pstone@cs.utexas.edu}
%
%
%
%
%
%

\begin{abstract}  


Curriculum learning in reinforcement learning is a training methodology that seeks to speed up learning of a difficult target task, by first training on a series of simpler tasks and transferring the knowledge acquired to the target task. Automatically choosing a sequence of such tasks (i.e.~a \emph{curriculum}) is an open problem that has been the subject of much recent work in this area. 
In this paper, we build upon a recent method for curriculum design, which formulates the curriculum sequencing problem as a Markov Decision Process. We extend this model to handle multiple transfer learning algorithms, and show for the first time that a \emph{curriculum policy} over this MDP can be learned from experience. We explore various representations that make this possible, and evaluate our approach by learning curriculum policies for multiple agents in two different domains. The results show that our method produces curricula that can train agents to perform on a target task as fast or faster than existing methods.  



\end{abstract}

%


\maketitle


\section{Introduction}



Over the past two decades, transfer learning \cite{Lazaric11,Taylor09} is one of several lines of research that have sought to increase the efficiency of training reinforcement learning agents. In transfer learning, agents train on simple \emph{source} tasks, and transfer knowledge acquired to improve learning on a more difficult \emph{target} task. Typically, this has been a one-shot process, where information is transferred from one or more sources directly to the target task. However, as the problems we task reinforcement learning agents with become ever more complex, it may be beneficial (and even necessary) to gradually acquire skills over multiple tasks \emph{in sequence}, where each subsequent task builds upon knowledge gained in a previous task. This insight is the basis for \emph{curriculum learning} \cite{Bengio09, Narvekar16}.

The goal of curriculum learning is to design a sequence of source tasks (i.e.~a curriculum) for an agent to train on, such that after training on that sequence, learning speed or performance on a \emph{target} task is improved. Automatically designing a curriculum is an open problem that has only recently begun to be examined \cite{Florensa17, Held18, Riedmiller18, Silva18, Narvekar17, Svetlik17}.
One recent approach \cite{Narvekar17} proposed formulating the selection of tasks using a (meta-level) curriculum Markov Decision Process (MDP). A policy over this MDP, called a curriculum policy, maps from the current knowledge of an RL agent to the task it should learn next. However they did not demonstrate whether the curriculum policy could actually be learned.
Instead, they proposed an algorithm to approximate a single execution of the curriculum policy, corresponding to an individual curriculum.

Until now, it was not known if curriculum policies could be learned: that is, whether it is possible to find a representation that is both compact enough and generalizable enough to facilitate learning. Our main contribution is to demonstrate that curriculum policies can indeed be learned, and we explore various representations that make this possible. In addition, we generalize the curriculum MDP model proposed by \citet{Narvekar17} to handle different kinds of transfer learning algorithms.
Finally, we empirically show that the curricula produced by our method are at least as good as, or better than those produced by two existing curriculum methods on two different domains. We also demonstrate that curriculum policies can be learned for agents with different state and action spaces, agents that use different transfer learning algorithms, and different representations for the curriculum MDP.

\section{Background}
\label{sec:Background}

Our work builds upon multiple ideas in reinforcement learning, transfer learning, and curriculum learning. In this section, we briefly describe background on each of these areas.

\subsection{Reinforcement Learning}
\label{sec:MDP}

We model an agent's interaction with its environment as an episodic Markov Decision Process (MDP). 
An episodic MDP $M$ is a 6-tuple $(\mathcal{S}, \mathcal{A}, p, r, S_0, S_f)$, where $\mathcal{S}$ is the set of states,
 $\mathcal{A}$ is the set of actions,
 $p(s, a, s')$ is a transition function that gives the probability of transitioning to state $s'$ after taking action $a$ in state $s$, and $r(s, a)$ is a reward function that gives the immediate reward for taking action $a$ in state $s$. In addition, we will use $S_0$ to denote the initial state distribution, and $S_f$ to represent the set of terminal states. 
 
At each time step $t$, the agent observes its state and chooses an action according to its \emph{policy} $\pi: \mathcal{S} \mapsto \mathcal{A}$. The goal of the agent is to learn an \emph{optimal policy} $\pi^*$, which maximizes 
 its expected 
\emph{return} 
(the cumulative sum of rewards) until the episode ends. 
There are two main classes of methods for learning $\pi^*$: value-function-based approaches and policy-search-based approaches. In this paper, we will primarily consider value-function-based approaches. We expect adapting our methods to policy-search-based approaches to be a relatively straightforward extension.

Value-function-based approaches first learn a value function $V_\pi(s)$, which gives the expected return achievable from state $s$ by following policy $\pi$. When a model is not known, an action-value function $Q_\pi(s,a)$ can be learned instead, 
which gives the expected return for taking action $a$ in state $s$ and following $\pi$ thereafter, using an algorithm such as SARSA or Q-learning \cite{Sutton98}. The optimal policy $\pi^*$ is derived by acting greedily with respect to $Q^*(s,a)$. 

When the state space is large or continous, it is not practical to maintain a separate entry in a lookup table for each state-action value. Instead, the state is represented by a set of state variables, and the 
value function is approximated as a function of \emph{state features} $\bm{\phi}(s)$ derived from the state variables of $s$, and a weight vector $\bm{\theta}$, 
 which instantiates the function approximation.

There are many options for extracting state features $\bm{\phi}(s)$ from the state variables of $s$. 
One option which we use in this paper is tile coding \cite{Sutton98}, which overlays grid-like \emph{tilings} over different subsets of state variables. The value of the state variables determine which \emph{tiles} in each tiling are activated, and the activated tiles form the features $\bm{\phi}(s)$. Each tile is associated with a weight which contributes equally towards the output of the function approximation. Typically, multiple overlapping tilings are used, which allows the approximator to generalize. 


There are also different types of functions we can use to represent the value or action-value funtion. Linear function approximators calculate the value as the inner product between the weight vector and feature vector: $\bm{\theta} \cdot \bm{\phi}(s)$. Similarly, the action-value can be calculated as $\bm{\theta} \cdot \bm{\phi}(s,a)$. \footnote{When it is clear from context, we will abbreviate $\bm{\phi}(s,a)$ and $\bm{\phi}(s)$ as $\bm{\phi}$.} Non-linear function approximators such as neural networks are also possible, and have become increasingly common in recent years. The goal during learning is to find weights $\bm{\theta}$, such that the policy $\pi_\theta(s)$ derived from its associated value function is optimal. 

\subsection{Transfer Learning}



When the target environment or task is too difficult to make progress on, learning can be accelerated by training on one or more source tasks, and transferring the knowledge acquired to the target task. This idea is the basis for transfer learning \cite{Taylor09,Lazaric11}. Many different techniques exist to transfer knowledge from one task to another. In this paper, we will use value function transfer \cite{Taylor07} and transfer via reward shaping \cite{Svetlik17}.

In value function transfer, the parameters of an action-value 
function $Q_{\mbox{\textit{source}}}(s, a)$ learned
in a prespecified source task are used to initialize the action-value function in the target task $Q_{\mbox{\textit{target}}}(s, a)$. Doing so provides an initialization bias that can allow an agent to explore more efficiently in the target task.     

In transfer via reward shaping, the reward function in the target task is augmented by adding an additional shaping reward $f$, that is derived from the source tasks. Thus, the new reward function becomes: 
\begin{equation}
r'(s,a,s') = r(s,a,s') + f(s,a,s')
\end{equation}

We use potential-based advice \cite{Wiewiora03}, which restricts the form of $f$ to be a difference of potential functions:

\begin{equation}
f(s,a,s') = \Phi(s',\pi(s')) - \Phi(s,a)
\end{equation}

where $\Phi$ is a potential function. Choosing shaping rewards of this form guarantees that adding $f$ to the reward does not change the optimal policy \cite{Ng99}. Following the work of \citet{Svetlik17}, we use the value function learned in a source task as the potential function: $\Phi(s,a) = Q_{\mbox{\textit{source}}}(s, a)$. When multiple source tasks are present, as will be the case in curriculum learning, the potential function is composed as the sum of value functions from the set of sources $\mathcal{X}$:
\begin{equation}
\Phi(s,a) = \sum_{i \in \mathcal{X}} Q_i(s,a)
\end{equation}

\subsection{Curriculum Learning}

Curriculum learning is an extension of transfer learning, where 
the goal is to automatically design and choose a full sequence of tasks (i.e.~a \emph{curriculum}) $M_1, M_2, \ldots M_t$ for an agent to train on, such that learning speed or performance on a target task $M_t$ is improved. Transfer learning is leveraged to transfer information between each pair of tasks in this sequence.

Our work builds upon the model proposed by \citet{Narvekar17}, which formulates curriculum generation as an interaction between two agents acting in two different MDPs. 
One is a \emph{learning agent} that is trying to solve a specific target task MDP $M_t$, as is the standard case in reinforcement learning. The second is a \emph{curriculum agent}, which interacts in a second, higher level \emph{curriculum MDP}, and whose goal is to sequence tasks $M$ for the learning agent. The way the process unfolds is as follows: the learning agent starts with some initial policy -- this is represented as the initial state of the curriculum agent. The curriculum agent selects an action, which corresponds to a task to learn. The learning agent interacts with that task, and updates its policy as a result of learning, which corresponds to a tranistion in the curriculum agent's state. Learning a task also returns a reward, which is the cost of learning that task. The process terminates once the learning agent learns a policy that can achieve a desired performance threshold on the target task.

This process was defined formally as follows (the superscript $C$ denotes elements of the curriculum MDP; the superscript is dropped when referring to the learning agent trying to solve the task): 

\textbf{Definition 1:} A \emph{curriculum MDP (CMDP)} $M^C$ is a 6-tuple \newline
$(\mathcal{S}^C, \mathcal{A}^C, p^C, r^C, S_0^C, S_f^C)$, where:

\begin{description}
\item[State Space] The set of states $\mathcal{S}^C$ consist of the set of all policies $\pi$ the learning agent can represent, in a form that is executable on the target task. 
For example, the initial state $S^C_0$ could be the uniform random policy. 
The terminal states $S^C_f$ are states whose policies achieve a return of at least some desired performance threshold $\delta$ on the target task. 

\item[Action Space]
The set of actions $\mathcal{A}^C$, are the prespecified set of tasks a learning agent can train on.  

\item[Transition Function]
The transition function $p^C(s^C, a^C, s'^C)$ 
gives the probability that $s'^C$ is the learning agent's policy after training on $a^C$  and starting with policy $s^C$.

\item[Reward Function]
The reward function $r^C(s^C, a^C)$ is the negative of the time (measured e.g.\ in experience samples or wall clock time) needed to learn task $a^C$ starting from policy $s^C$. 
\end{description}

A policy $\pi^C: \mathcal{S}^C \mapsto \mathcal{A}^C$ on a CMDP specifies which task to train on given a learning agent policy $s^C$. Executing $\pi^C$ for a particular learning agent produces a curriculum. Learning a full policy over a CMDP can be very difficult, due to stochasticity in the learning algorithm (which leads to stochasticity in the CMDP transition function), a very large and continuous state space, and the high cost of taking a CMDP action. 
Thus, past work on explicit curriculum generation has tried to find traces of specific curricula using approximations and heuristics, rather than learning a full CMDP policy \cite{Narvekar17,Svetlik17}. In this work, we explore the challenges involved in learning $\pi^{C^*}$.





\section{Learning Curriculum Policies}
\label{sec:learnCMDP}


Before discussing how to learn a curriculum policy, we first briefly extend the definition of a curriculum MDP. A shortcoming of the previous definition is that it assumes the underlying transfer learning mechanism is value function or policy transfer. Intuitively, the state space of a CMDP represents different states of knowledge. A transition between states reflects the change in knowledge from training on a task and \emph{transferring/incorporating} the information acquired. In value function transfer, the knowledge learned from a task is represented by the value function of the agent itself. However, for transfer via reward shaping, knowledge is represented in terms of a potential-based shaping reward.

Thus, the CMDP state space and transition function are directly related with the transfer learning algorithm being used. The goal of the agent is to reach a state of knowledge that allows solving the target task in the least amount of time. 
Therefore, for an agent that uses reward shaping, the CMDP state is represented as a set of potential functions, derived from the value functions of source tasks already learned. The goal is to find a CMDP state whose sum of potential functions creates a shaping reward that allows learning the target task as fast as possible.

\subsection*{Representing CMDP State Space}

\label{sec:whitebox}



We now detail how to represent the CMDP state to facilitate learning of curriculum policies. 
Recall that in the standard reinforcement learning setting, the agent perceives its state as a set of raw state variables. These are typically used to extract basis features $\bm{\phi}(\mathbf{s})$, which transform the state variables into a space more suitable for learning and for use in function approximation. 
Given these features and a functional form, the goal is to learn weights $\bm{\theta}$ for the value function or policy. We introduce an analagous process for curriculum design agents acting in CMDPs. 
We will ground the discussion assuming the learning agent uses value function transfer. However, the idea is easily applied to the reward shaping setting by noting that the potential-based reward, like the value function, can be expressed as a function of state features and weights.  


The first question is how to represent the raw state variables $s^C$ of the CMDP state space. The representation chosen must be able to represent \emph{any} policy the underlying learning agent can represent. 
Assuming the learning agent derives its policy from an action-value function $Q_{\bm{\theta}}(s,a)$, the form of the function -- in particular, the way values are calculated from $\bm{\phi}(s,a)$ and $\bm{\theta}$ (for example, the architecture of a neural network) -- determines the class of policies that can be represented. The functional form of $Q_{\bm{\theta}}(s,a)$ and how learning agent features $\bm{\phi}$ are extracted are fixed. Thus, it is specific values of the weight vector $\bm{\theta}$ that actually instantiates a policy in this class.
Therefore, it follows that we can represent the state variables for a particular CMDP state $s^C$ using the instantiated vector of learning agent weights $\bm{\theta}$. 
\begin{equation}
s^C = \bm{\theta}
\end{equation}
Different instantiations of $\bm{\theta}$ correspond to different CMDP states. Typically, these weights $\bm{\theta}$ will take on continuous values. Therefore, in order to learn a CMDP action-value function $Q^C_{\bm{\theta}^C}(s^C, a^C)$, 
 it will be necessary to do some kind of function approximation.
While it is possible to directly use the raw $\bm{\theta}$ as features for function approximation in the CMDP, learning may be more efficient in an alternative basis space. Thus, it may be beneficial to extract \emph{CMDP basis features} $\bm{\phi}^C(s^C, a^C)$, mirroring what is done in the standard MDP setting. For example, with linear value function approximation, $Q^C_{\bm{\theta}^C}(s^C, a^C) = \bm{\theta}^{C}\cdot\bm{\phi}^C(s^C,a^C)$. 
The goal then is to learn the weights $\bm{\theta}^C$ for the CMDP's value function. Any standard RL algorithm can be used to do this.

The questions that remain are: (1) how to convert raw CMDP state variables to CMDP basis features, i.e.~the form of $\bm{\phi}^C(s^C,a^C)$; and (2) what kind of functional form to use to represent the function approximation. 
The best way to do these will vary by domain. 
However, in the next 2 subsections, we provide specific examples and guidelines for representations and function approximations that can apply across a broad class of domains. 

\subsubsection*{\textbf{Discrete State Representations}}
\label{sec:discretecase}

First we propose one specific way of extracting CMDP state features 
and performing function approximation, that can be applied when the parameters $\bm{\theta}$ are tied to specific states, as is common in tabular reinforcement learning.

Assume again the learning agent learns an action-value function $Q_{\bm{\theta}}(s,a)$, for each state-action pair in the task. We can represent $Q$ as a linear function of ``one-hot" features $\bm{\phi}(s,a)$ and their associated weights $\bm{\theta}$:
\begin{equation}
Q_{\bm{\theta}}(s,a) = \bm{\theta} \cdot \bm{\phi}(s,a) 
\end{equation}

In other words, all the action-values are stored in $\bm{\theta}$, and $\bm{\phi}(s,a)$ is a one-hot vector used to select the activated action-value from $\bm{\theta}$. Our approach for designing $\bm{\phi}^C$ is to utilize tile coding over subsets of action-values in $\bm{\theta}$. 
Specifically, the idea is to create a separate tiling for each primitive state $s$ in the domain. Each such tiling will be defined over the action-values in $\bm{\theta}$ associated with state $s$. Thus, this creates $|\mathcal{S}|$ tiling groups, where each group is defined over $|\mathcal{A}|$ CMDP state variables (i.e.~action-values). To create the feature space, multiple overlapping tilings are laid over each group.





Since action-values can take a large range of values, we suggest normalizing the action-values within each tiling. Thus, each tiling is over the relative preferences of the different actions in a state. The entire CMDP basis state is the concatenation of all of these tiled features.
The effect of this approach is that when computing the value of a CMDP state $s^C$, the policy for each primitive state contributes equally towards the total value. Two CMDP states 
will 
be ``closer" in representation space the more $\bm{\phi}^C$ activates the same tiles -- which will happen if they have similar action preferences for primitive states in their task state spaces.







\subsubsection*{\textbf{Continuous State Representations}}
\label{sec:continuouscase}

The representation problem is harder in the continuous case, since each parameter $\theta_i$ is not local to a state, and we cannot 
use a state-by-state approach to create a basis feature space. 
In principle, any continuous feature extraction and function approximation scheme can form the basis of $\bm{\phi}^C$ (tile coding, neural nets, etc.). 
We offer 2 guidelines that we found useful in defining successful $\bm{\phi}^C$  representations in our experiments.

The first is that the precise form of $\bm{\phi}^C$ should be informed by the domain and the structure of the learning agent's function approximation. The discrete case discussed previously is a special case of this setting. 
In the discrete case, aggregating action-values in a state-by-state basis could be thought of as exploiting the structure and what we know about 
 the parameter vector $\bm{\theta}$: namely, that it consists of action values that share states. Depending on the function approximation used by the learning agent, it may be possible to draw similar insights to design $\bm{\phi}^C$.


The second guideline for creating $\bm{\phi}^C$ is to capture the \emph{relative} effect of each $\theta_i$ on different action preferences. In the discrete case, this was done by normalizing the action values within each state to create preferences. However, since in general parameters may not be local to a state, the normalization needs to be done directly on the parameter values. In other words, we need to think about how each parameter $\theta_i$ affects the policy as a whole over all states, and how each parameter $\theta_i$ relates to another.
If the parameters $\bm{\theta}$ are not related, one option would be to create a separate tiling over each parameter, and normalize over all the parameter values. 
We will demonstrate a specific example of creating $\bm{\phi}^C$ for the continuous case in the experiments (Sections \ref{sec:whiteboxEx} and \ref{sec:pacman_cmdpstate}).



%


\section{Experimental Results}
\label{sec:experiments}


We evaluated learning curriculum policies for agents on a grid world domain \cite{Narvekar17} as well as a Ms.~Pac-Man domain \cite{14ConnectionScience-Taylor}. These domains were selected because they allow us to compare to previous methods; 
test our approach using different agent representations, different transfer learning algorithms, and different CMDP representations; and test its scalability to a more complex setting.

We will show the results as \emph{CMDP learning curves}. The x-axis on these learning curves are over \emph{CMDP episodes}. 
Each CMDP episode represents an execution of the current curriculum policy for the agent. Thus, multiple tasks are selected over the course of a single CMDP episode, with each task taking a varying number of steps/episodes, which contributes to the cost on the y-axis. Tasks are selected until the desired performance can be achieved in the target task, at which point the CMDP episode is terminated. In short, the curves show how long it would take to achieve a certain performance threshold on the target task following a curriculum, where the curriculum is represented by the CMDP policy, which is being learned over time.

We compare curriculum policies learned for each agent to two static curricula. The first is the baseline \emph{no curriculum} policy. In this case, on each episode, the agent learns tabula rasa directly on the target task. The flat line plotted represents the average time needed to learn the target task directly. Note that the line is flat because the ``curriculum" is fixed and does not change over time. The second is a curriculum produced by following an existing curriculum algorithm (from \cite{Narvekar17} for the gridworld and from \cite{Svetlik17} for Ms.~Pac-Man, to compare with past work). 
We also compare to a naive learning-based approach, which represents CMDP states using a list of all tasks learned by the learning agent. For example, the start state is the empty list. Upon learning a task $M_1$, the CMDP agent transitions to a new state $[M_1]$. If the CMDP agent subsequently selects task $M_t$, the resulting state is $[M_1, M_t]$. Note that this representation is a cruder approximation of the underlying process, as learning 2 different tasks that impart the same knowledge will lead to 2 different states under this representation. In order to deal with the combinatorial explosion of the size of the state space with this naive representation, we limit the number of tasks that can be used as sources in the curriculum to a constant (between 1 and 3 in our experiments), and force the selection of the target task after. 

Hyperparameters for the learning agents were chosen using previously reported results in the respective domains. Hyperparameters for the CMDP agents were set as described in Sections \ref{sec:gridworld_cmdp_desc} and \ref{sec:pacman_cmdp_desc}. These were not extensively optimized. 

\subsection{Gridworld Experiments}

\begin{figure}[tbp]
\begin{center}
\includegraphics[height=2.8cm]{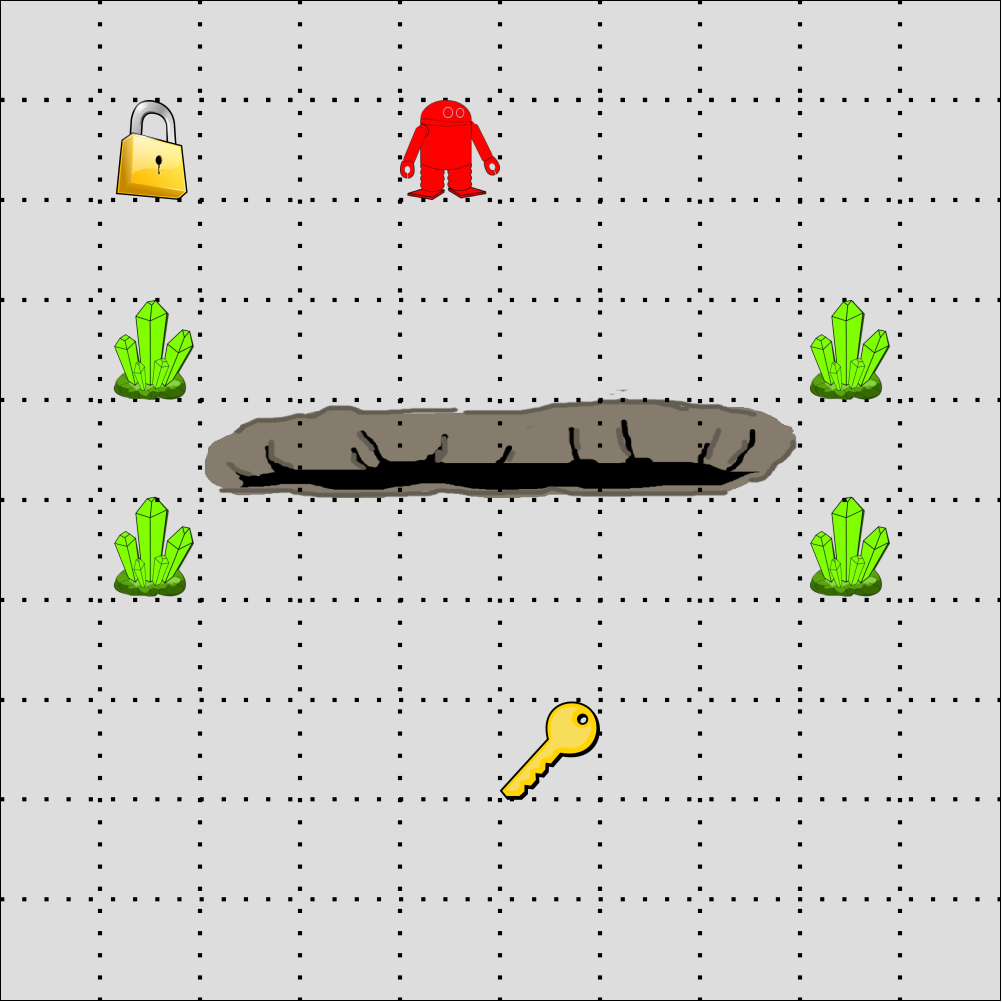} 
\end{center}
\caption{ Grid world target task.}
\label{domain}
\end{figure}

First we examine learning curriculum policies for 3 learning agents that have different state and action spaces \cite{Narvekar17}, but use the same transfer learning algorithm (value function transfer), in a grid world domain (see Figure \ref{domain}). In addition, we compare the effect of two different types of representations for the CMDP state. 
The first CMDP representation is based on the finite state space representation discussed earlier, while the second CMDP representation is created directly from $\bm{\theta}$ without using an intermediary state-based action-value representation. In the next sections, we describe the domain and learning agents, followed by the representations for the CMDP state space and their effects on learning curriculum policies.


\subsubsection{Domain Description}

The gridworld consists of a room with 4 different types of objects. \emph{Keys} are objects the agent can pickup by executing a \emph{pickup} action. These are used to unlock \emph{locks}, which can depend on one or more keys, by executing an \emph{unlock} action. \emph{Pits} are obstacles that terminate the episode upon contact. 
Finally, \emph{beacons} are landmarks placed on the 4 corners of a pit. 

The goal of the agent is to traverse the world and unlock all the locks. To do so, at each step, the agent can move in one of the 4 cardinal directions, execute a pickup action, or an unlock action. Successfully picking up keys give a reward of +500, unlocking a lock rewards +1000, falling into a pit ends the episode with a reward of -200, and a constant step penalty of -10 is applied for all other actions.


\subsubsection{Learning Agent Descriptions}
\label{sec:learningagents}


We created 3 different learning agents that have varied sensing and action capabilities, based on the agents presented by \citet{Narvekar17}. 
Creating these specific agents allows us to show that our approach works regardless of the state and action representation used by the learning agents, and also allows us to compare with the results of \citet{Narvekar17}. 
We refer the reader there for full details of the agents, but recap the main elements and differences here for completeness. 

The first agent, the \emph{basic agent}, has 16 sensors, grouped into 4 on each side. The first sensor in each quadruple measures the Euclidean distance to the closest key from that side, the second the distance to the closest lock, the third the distance to the closest beacon, and the fourth detects whether there is a pit adjacent to the agent in that direction. 
An additional sensor indicates whether all keys in the room have been picked up, referred to as the \emph{noKey} sensor. 
The agent used Sarsa$(\lambda)$ with $\epsilon$-greedy action selection, value function transfer for transfer learning, and CMAC tile coding with linear function approximation,  
where tile widths were set to 1. 

For the basic agent, we created two tilings: one over the 13 percepts from the key, beacon, pit, and noKey sensors, and another over the 13 percepts from the lock, beacon, pit, and noKey sensors. These tilings formed $\bm{\phi}(s)$ for the basic agent.  
The exploration rate $\epsilon$ was set to 0.1, eligibility trace parameter $\lambda$ to 0.9, and learning rate $\alpha$ to 0.1 (these values match those reported in \cite{Narvekar17}).

The second, \emph{action-dependent agent}, has the same sensors as the basic agent, but they are tiled differently, leading to a different $\bm{\phi}(s)$: one tiling is over the lock, pit, and noKey features; a second is over the key, pit, and noKey features; and a third is over the beacon and pit features. In addition, unlike the basic agent, the state representation is action-dependent. That is, when considering the \emph{move right} action, the agent's feature vector uses values only from the right side sensors. 
The weights in the tilings are shared, so that the same set of weights is used for the state in each of the directions. 




 


Finally, the \emph{rope agent} is like the basic agent, except that it has 4 additional actions, which are to use a rope in one of the four directions. Doing so opens a path across a pit if one is present, and incurs the step cost of -10. 

\begin{figure*}[t]
\begin{center}$
\begin{array}{ccc}
\includegraphics[height=4.cm]{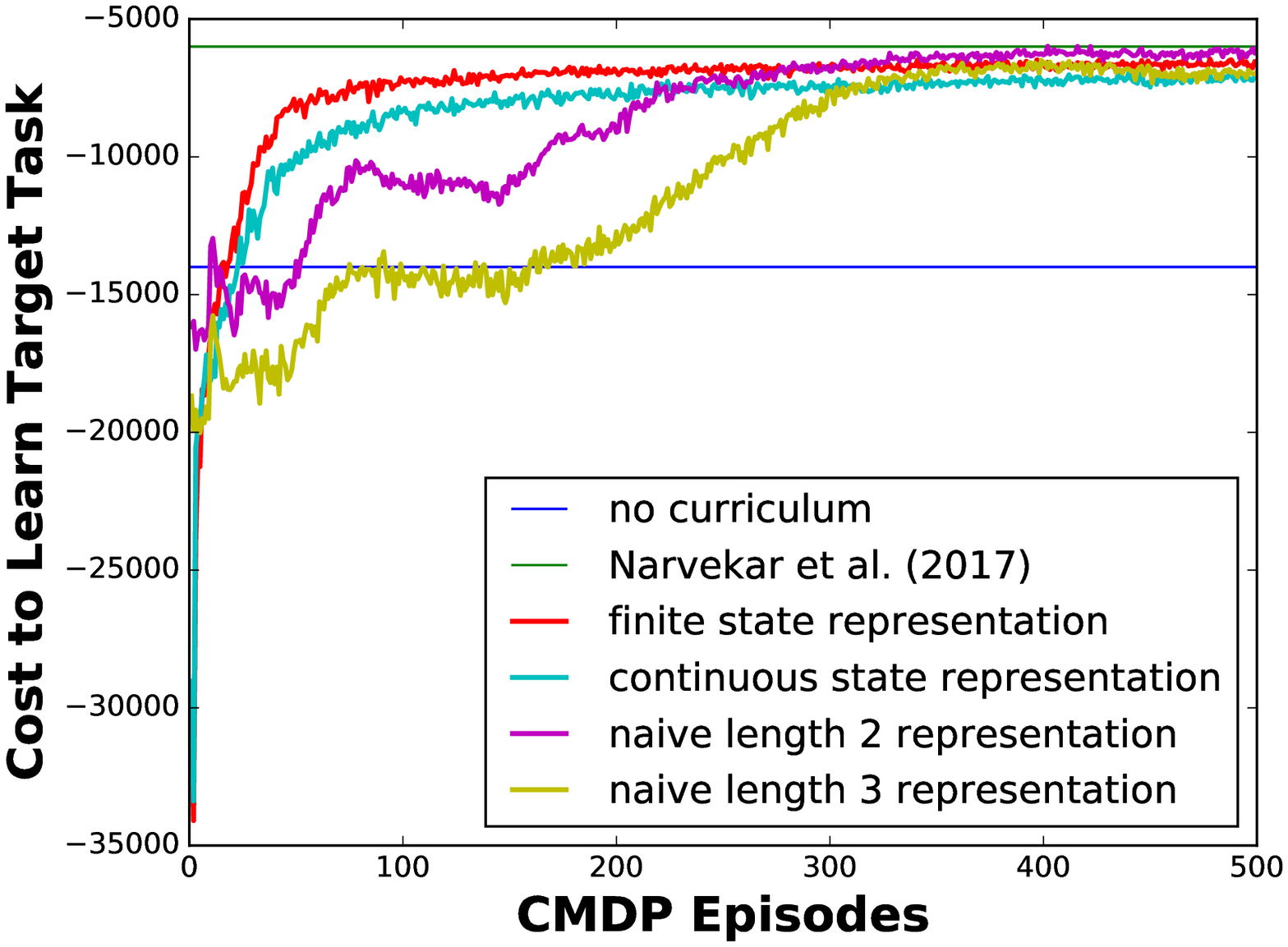} & \includegraphics[height=4.cm]{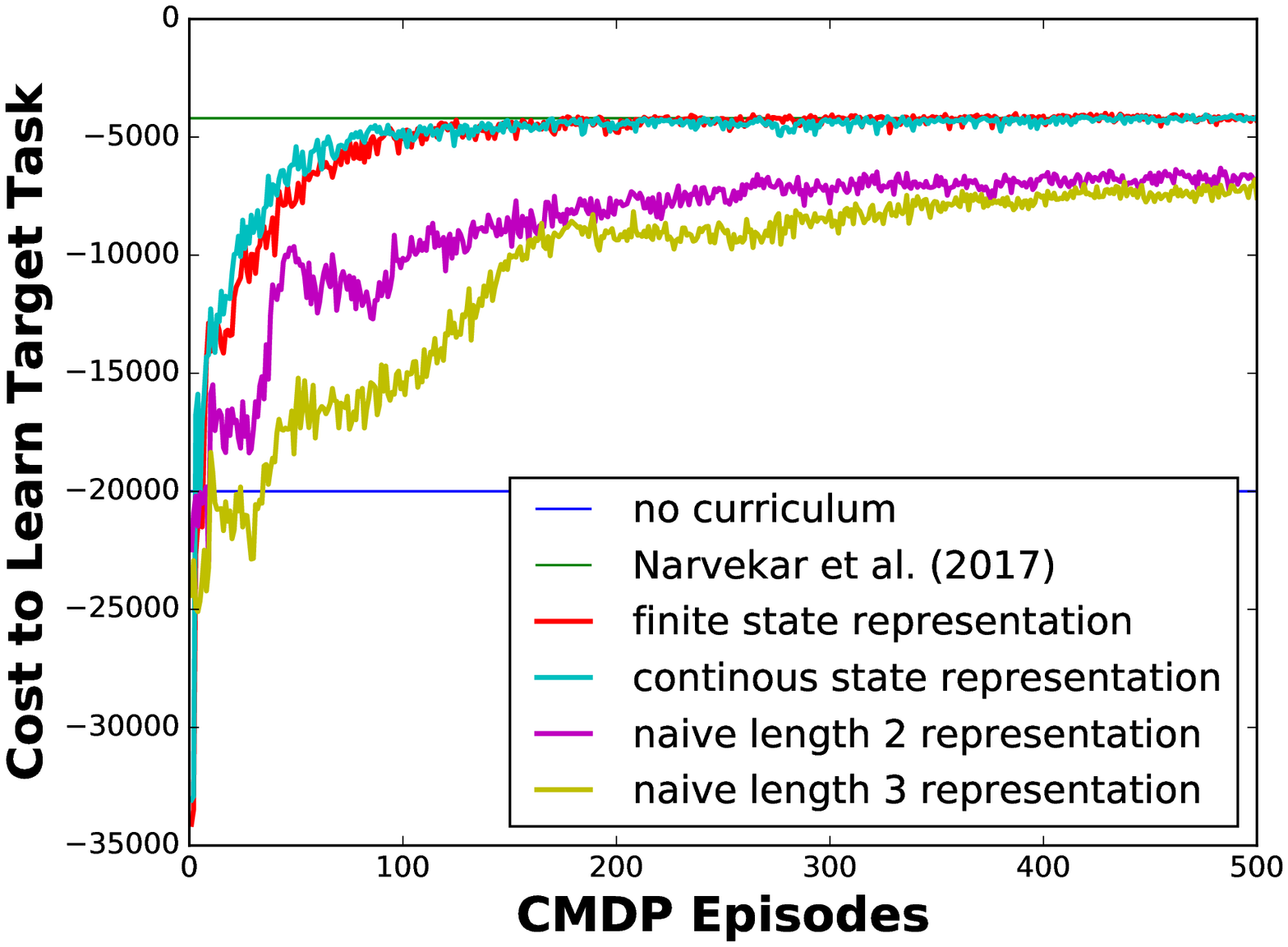} & \includegraphics[height=4.cm]{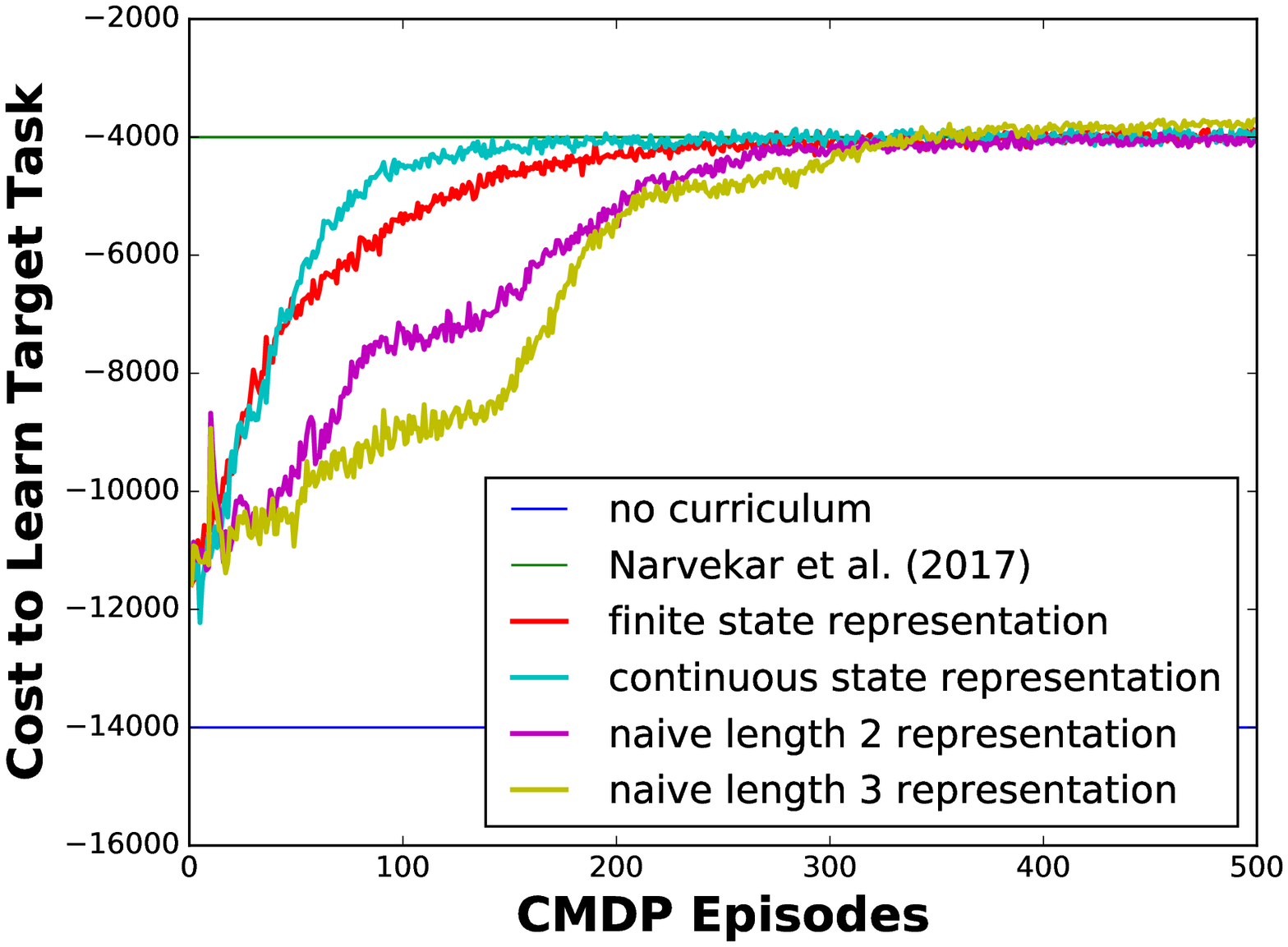} \\
(a) & (b) & (c)
\end{array}$
\end{center}
\caption{CMDP learning curves for the $(a)$ basic agent, $(b)$ action-dependent agent, and $(c)$ rope agent using different curriculum design approaches and CMDP state space representations. The y-axis represents the cost (i.e.~negative of the time needed) to reach a performance of 700 on the target task, following the curriculum policy at episode X. All curves are averaged over 500 runs. }
\label{cmdpLearningResults}
\end{figure*}

\subsubsection{CMDP Description}
\label{sec:gridworld_cmdp_desc}

We defined our curriculum MDP as follows:

\textbf{State space.} The start state $S^C_0$ was derived from an untrained, uniformly initialized learning agent. The set of terminal states $S^C_f$ were all states where the learning agent's policy allowed it to achieve a return of at least 700 on the target task. This performance threshold was the maximum that all the agents could achieve after training to convergence on the target task. Representations used for the CMDP state space are described in the next section. 
 
\textbf{Action space.} Source tasks were created using the TaskSimplification and OptionSubGoals heuristics \cite{Narvekar16}. These heuristics create source tasks by simplifying the domain, for example by reducing the size of the grid or the number of keys, locks, and pits, and by changing the goal of the task to be picking up keys. A total of 10 different tasks were created, and with the target task, these formed the action space $\mathcal{A}^C$ of the CMDP agent. The properties of these source tasks are summarized in Table \ref{gridworld_tasks}.

\textbf{Transition function.} 
The (unknown) transition function is stochastic, describing how learning a task changes a learning agent's policy.

\textbf{Reward function.}
The environment returns a reward $r^C(s^C, a^C)$ as the negative of the time needed to learn task $a^C$ from state $s^C$. A task is considered learned once the policy ceases to change for 10 episodes. Time is measured using game steps. 

Learning on the CMDP was done using Sarsa$(\lambda)$ with $\epsilon = 0.001$, $\lambda = 0.9$, and $\alpha	 = 0.1$. 

\begin{table}
\scriptsize
\begin{center}
\begin{tabular}{|c|c|c|c|c|c|}

\hline
\textbf{Task Num} & \textbf{Grid Size} & \textbf{Num Keys} & \textbf{Num Locks} & \textbf{Pit Present} & \textbf{Rope Required} \\
\hline
\hline
1 & 5x5 & 1 & 0 & No & No \\
2 & 10x10 & 1 & 0 & No & No\\
3 & 5x5 & 0 & 1 & No & No\\
4 & 10x10 & 0 & 1 & No & No \\
5 & 7x1 & 1 & 0 & Yes & Yes \\
6 & 7x6 & 1 & 0 & Yes & Yes \\
7 & 7x1 & 0 & 1 & Yes & Yes \\
8 & 7x6 & 0 & 1 & Yes & Yes \\ 
9 & 7x7 & 1 & 0 & Yes & No \\
10 & 7x7 & 0 & 1 & Yes & No \\
Target & 10x10 & 1 & 1 & Yes & No \\
\hline

\end{tabular}

\end{center}
\caption{Properties of tasks in the gridworld experiments. ``Rope required" indicates tasks where a pit blocks direct paths from the agent to the goal, necessitating a rope action. When a lock is not present, the episode terminates when all keys are picked up.} 
\label{gridworld_tasks}
\end{table}


\subsubsection{CMDP State Space Representations}
\label{sec:whiteboxEx}



One of the main challenges addressed in this research is identifying a representation for the CMDP state space that is both generalizable and compact enough to enable efficient learning of a curriculum for a range of agents. To this end, we instantiated and evaluated two forms for $\bm{\phi}^C$. 

Recall that the learning agents use tile coding with linear function approximation. 
Here, $\bm{\phi}$ is a feature vector that indicates which tiles have been activated for state $s$ and action $a$, and $\bm{\theta}$ are the corresponding weights in each tile. These weights $\bm{\theta}$ form the raw CMDP state variables $s^C$. We discuss two different ways to construct $\bm{\phi}^C(\bm{\theta})$, which will convert the raw state variables into a CMDP basis feature space suitable for learning.



\textbf{Finite State Representation.}
The learning agents use Sarsa($\lambda$) with an \emph{egocentric} feature space. Thus, the parameters $\bm{\theta}$ learned are not action-values for each state. However, since the underlying domain has a fixed number of states, we can simulate the finite state representation case by moving the learning agent to each of the states in the target task and computing action values. 
Let this new parameter of weights be $\bm{\theta}'$. We can now utilize the procedure described in Section \ref{sec:learnCMDP} to create a CMDP feature space $\bm{\phi}^C(\bm{\theta')}$.

\textbf{Continuous State Representation.}
The above representation is only well-defined in environments with a discrete underlying state space.  We therefore also explore a CMDP representation that can apply in continuous domains by creating $\bm{\phi}^C$ directly from $\bm{\theta}$ without using an intermediary state-based action-value representation. 
Recall that the CMDP state variables $s^C = \bm{\theta}$ are the weights associated with all the tiles. 
Each of these tiles is part of a tiling group. For example, the basic and rope agents had 2 tiling groups over different subsets of its sensor percepts, while the action-dependent agent had 3 tiling groups. 
All tiles in a tiling group are related to each other. Thus there is an inherent structure to the parameters in the tiles.

However, forming a $\bm{\phi}^C$ tiling group over the weights of all the tiles associated with a $\bm{\phi}$ tiling would not generalize well, because it would require nearly identical action-preferences in every state to activate common tiles. Therefore, we created a separate tile group for each $\theta_i$. 
Since the weights $\bm{\theta}$ within each learning agent's tilings $\bm{\phi}$ are still correlated, we normalized the weights associated within each $\bm{\phi}$ tile group. 

\subsubsection{Results and Discussion}

We learned curriculum policies for all 3 learning agents 
 using both the finite and continuous state representations for the CMDP state space. 
The target task $M_t$ is shown in Figure \ref{domain}.
The corresponding CMDP learning curves are shown in Figures \ref{cmdpLearningResults}(a) - \ref{cmdpLearningResults}(c).
The results show that each agent successfully learned curriculum policies using both CMDP representations that were better than learning without a curriculum, and comparable to the curricula generated by previous work \cite{Narvekar17}. However, unlike this previous work, our approach does not require additional prior information about source tasks (such as task descriptors).
In addition, the results show that our approach is flexible to changes in representation for both the learning agents as well as the CMDP agent. 

\subsection{Ms.~Pac-Man Experiments}

\begin{figure}[t]
\begin{center}
\includegraphics[height=4cm]{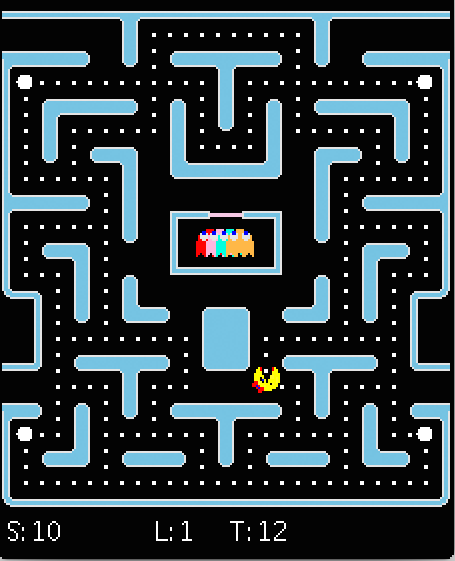} 
\end{center}
\caption{Ms.~Pac-Man target task}
\label{pacmanDomain}
\end{figure}

In the previous section, we demonstrated that CMDPs can be learned for agents with different actions and/or state representations. Another relevant way in which agents can differ is the algorithm by which they transfer knowledge from a source task to a target task.  Thus in this section, we evaluate the robustness of our approach to different underlying transfer learning methods, while simultaneously evaluating the scalability to a significantly more complex Ms.~Pac-Man domain (see Figure \ref{pacmanDomain}).
In particular, we examine the case when the learning agent stays the same, but uses 2 different types of transfer learning methods: value function transfer and reward shaping. The change in transfer algorithm affects both the CMDP state space representation, as well as the CMDP transition function, which we will describe in the following sections.

\subsubsection{Domain Description}

Our implementation is based on code released by \citet{14ConnectionScience-Taylor} and augmented by \citet{Svetlik17}. The goal of the Ms.~Pac-Man agent is to traverse a maze and accumulate points by eating edible objects such as food pellets, while avoiding ghosts. At each time step, Ms.~Pac-Man can move along one of the 4 cardinal directions (though not every action is available in every state). The agent receives a reward of 10 for eating a pill and 50 for a power pill. Eating a power pill temporarily makes the ghosts edible. Eating the first ghost gives a reward of 200; each subsequent ghost eaten multiplies this reward by a factor of 2. The game ends when all food pellets and power pills have been eaten, a ghost eats Ms.~Pac-Man, or when a time limit of 2000 game steps have occurred. 

While the domain is technically discrete, it has a combinatorially large state space. There are over a thousand positions in the target task maze, and the state consists of the locations of Ms.~Pac-Man, each food pellet, power pill, each of the 4 ghosts, the last move of each ghost, and whether each ghost is edible. Thus, it is essential for the Ms.~Pac-Man learning agent to use function approximation.

\subsubsection{Learning Agent Description}

We created a Ms.~Pac-Man learning agent using the low-asymptote feature set  described in \citet{14ConnectionScience-Taylor, Svetlik17}. The state space of the agent is represented by a set of action-dependent egocentric features, that count the fraction of pills, power pills, ghosts, and edible ghosts there are in each direction up to different ``depths." The depth is represented in terms of junctions, i.e.~locations in the maze where there are more than 2 possible actions. For example, the ghost feature for depth 1 would return the fraction of ghosts there are along one direction until the first junction. 
The features were used to learn a linear value function approximator.

The agent was trained using Sarsa$(\lambda)$, with $\epsilon = 0.05$, $\alpha = 0.001$, $\gamma = 0.999$, and $\lambda = 0.9$. 
 See the code by \citet{Svetlik17} for implementation details.
%



\subsubsection{CMDP Description}
\label{sec:pacman_cmdp_desc}

We defined our curriculum MDP as follows:

\textbf{State space.} As before, the start state $S_0^C$ was an untrained, randomly initialized learning agent. The set of terminal states $S^C_f$ were all states where the learning agent could achieve a return of at least 2000 on the target task.

\textbf{Action space.} We used the same 15 tasks used in the code release of \citet{Svetlik17} to form the action space $\mathcal{A}^C$. These tasks were formed by varying the type of maze as well as the number of pills, ghosts, and power pills. Their properties are summarized in Table \ref{pacman_tasks}.

\textbf{Transition function.} As before, the (unknown) transition function is stochastic, describing how Ms.~Pac-Man's value function or set of shaping potentials changes as a result of learning a task. 

\textbf{Reward function.} We measure the cost of learning a task in terms of the number of game steps needed. Following the experimental setup of \cite{Svetlik17}, a task is considered learned when at least 35\% of the maximum reward possible for that task can be achieved. The maximum reward for a task is calculated analytically by summing the points accrued for eating all the pills, and all the edible ghosts for each power pill. 

Learning on the CMDP was done using Sarsa$(\lambda)$ with $\epsilon = 0.001$, $\lambda = 0.9$, and $\alpha=0.05$.

\begin{table}

\scriptsize
\begin{center}
\begin{tabular}{|c|c|c|c|c|}
\hline
\textbf{Task Num} & \textbf{Num Junctions} & \textbf{Num Ghosts} & \textbf{Num Pills} & \textbf{Num Power Pills} \\
\hline
\hline
1 & 2 & 0 & 53 & 1 \\
2 & 2 & 1 & 65 & 2 \\
3 & 40 & 2 & 234 & 4 \\
4 & 36 & 4 & 240 & 4 \\
5 & 8 & 0 & 179 & 4 \\
6 & 8 & 2 & 179 & 4 \\
7 & 8 & 4 & 179 & 4 \\
8 & 13 & 2 & 209 & 4 \\
9 & 13 & 4 & 209 & 4 \\
10 & 13 & 0 & 209 & 4 \\
11 & 24 & 0 & 231 & 4 \\
12 & 24 & 2 & 231 & 4 \\
13 & 24 & 4 & 231 & 4 \\
14 & 24 & 4 & 231 & 4 \\
Target & 36 & 4 & 240 & 4 \\
\hline
\end{tabular}
\end{center}
\caption{Properties of source tasks in the Ms.~Pac-Man experiments. ``Num Junctions" indicates how many maze positions had 3 or more direction actions possible. Note that some tasks have similar properties; however, the layout of the maps in these tasks differed. Please see the code release from \citet{Svetlik17} for more details.}
\label{pacman_tasks}
\end{table}

\subsubsection{CMDP State Space Representations}
\label{sec:pacman_cmdpstate}

We consider 2 different CMDP state space represenations that result from the use of 2 different transfer learning algorithms.
In the value function transfer case, the raw CMDP state variables $s^C$ are the weights $\bm{\theta}$ of the Ms.~Pac-Man agent's linear function approximator. To create the CMDP space $\bm{\phi}^C$, we normalize $\bm{\theta}$ and use tile coding, creating a separate tiling over each $\theta_i$. In the reward shaping setting, each source task in the curriculum is associated with a potential function (derived from the value function). As multiple tasks are learned, the potentials are added together, and used to create a shaping reward (as done in \citet{Svetlik17}). Thus, the raw CMDP state variables are the summed weights of the potential functions. As in the value function case, we use tile coding to create a separate tiling over each potential weight feature to create the CMDP basis space.

\subsubsection{Results and Discussion}


\begin{figure*}[t]
\begin{center}$
\begin{array}{ccc}
\includegraphics[height=4cm]{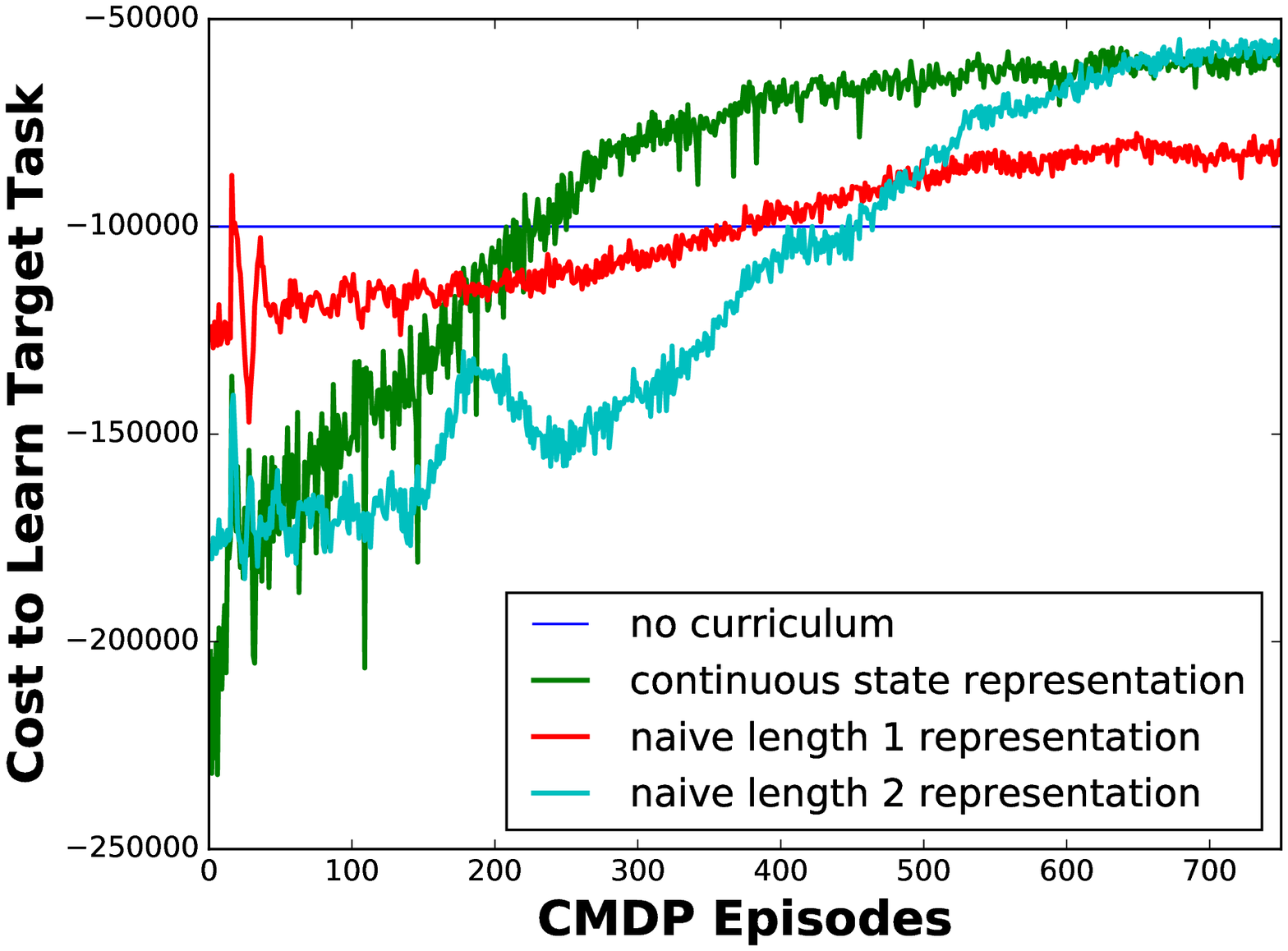} &  \includegraphics[height=4cm]{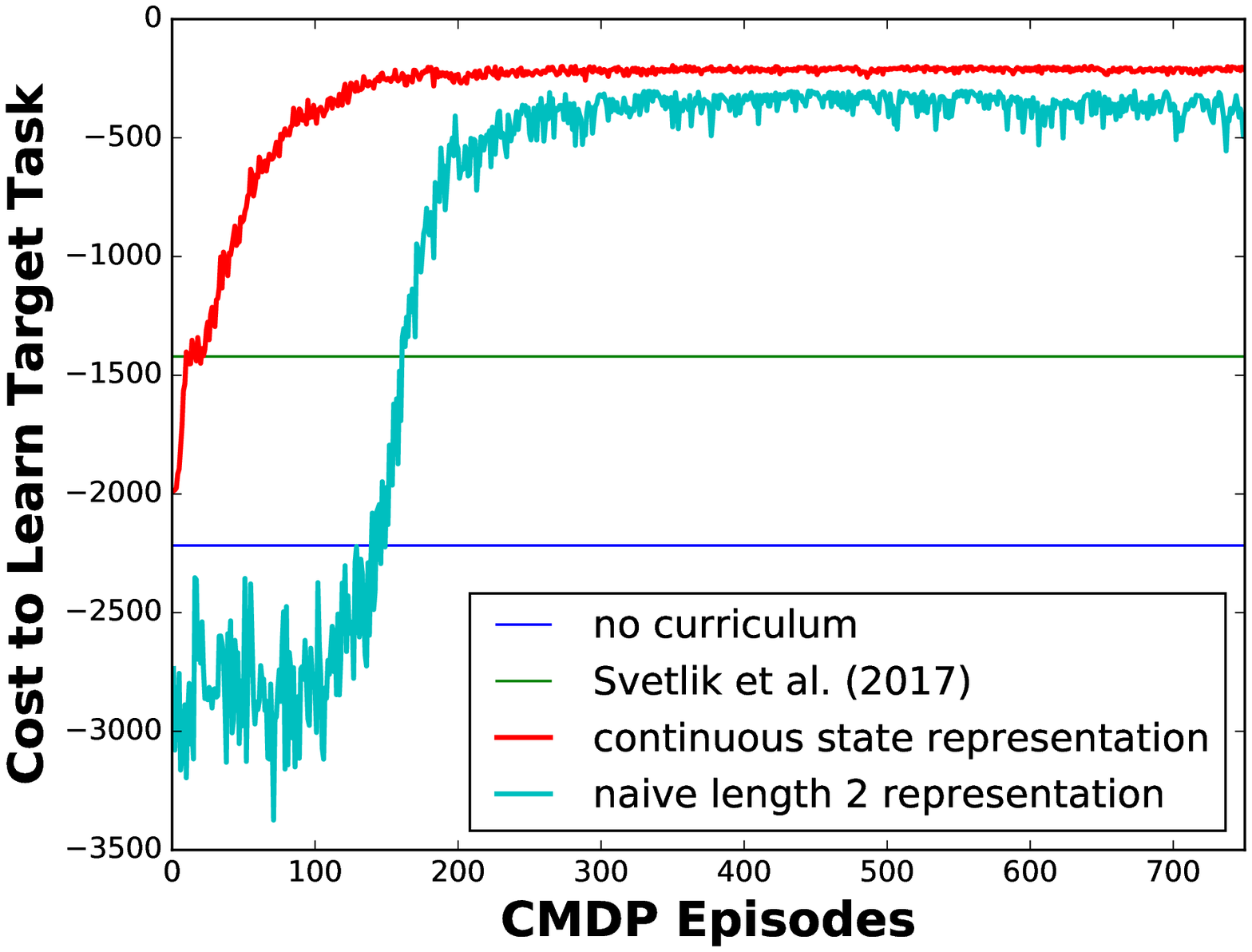} & \includegraphics[height=4cm]{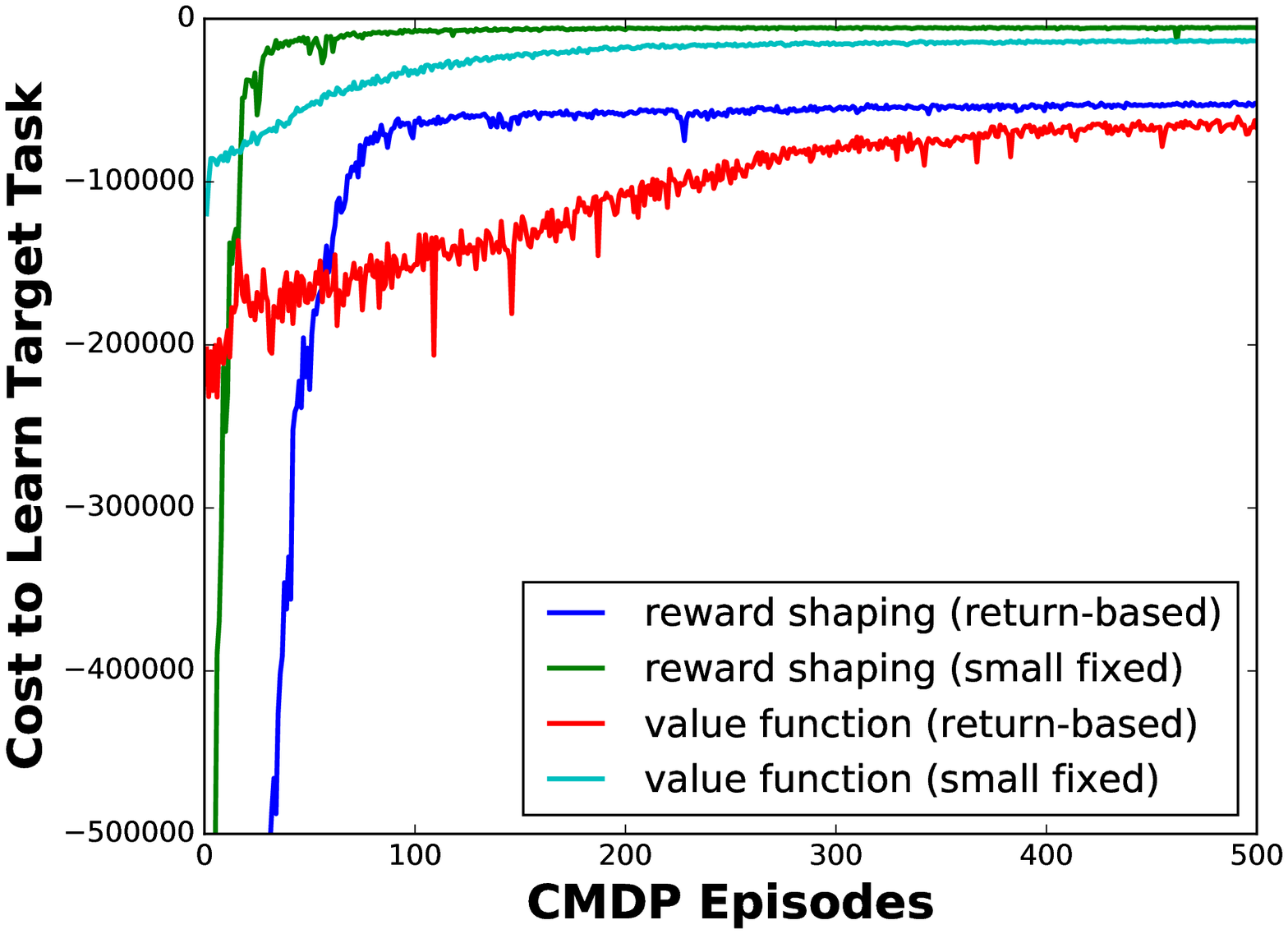} \\
(a) & (b) & (c)
\end{array}$
\end{center}
\caption{CMDP learning curves on the Ms.~Pac-Man target task, using (a) value function transfer, (b) transfer with reward shaping, and (c) a comparison between the continuous representations for value function transfer and reward shaping transfer, using different criteria to determine when to stop training on source tasks. All curves are averaged over 500 runs. Cost is measured in game steps for (a) \& (c), and episodes for (b).}
\label{pacmanAgent}
\end{figure*}

Figure \ref{pacmanAgent}(a) shows CMDP learning curves for Ms.~Pac-Man using value function transfer and Figure \ref{pacmanAgent}(b) shows the curves using transfer with reward shaping. The results again clearly demonstrate that curriculum policies can be learned, and that such policies are more useful than training directly on the target task. They also show that the approach is adjustable to different types of transfer learning algorithms. In addition, we compared the reward shaping approach with that of \citet{Svetlik17}, who also use reward shaping for transfer in their curriculum algorithm, and found that a much better curriculum is possible in this more complex domain.\footnote{Our results are based on a reproduction of their experiments using their publicly released code. Interestingly, we also get slightly better results for their method than they report in their paper. We measure cost in episodes for this experiment only to facilitate comparison to their work.} 

Finally, we also study the effect of the hyperparameter that controls when to finish training on a source task. For the previous two experiments in Ms.~Pac-Man, training on a source was stopped after 35\% of the max possible return in the task was achieved, to replicate the experimental conditions of \citet{Svetlik17}. Since their approach precomputes a curriculum and does not model the state of the learning agent's progress, this termination condition must be carefully chosen to ensure something can learned in each source task. 
In contrast, with our approach, we can train on source tasks for an arbitrarily small amount of time, as the curriculum policy can learn to reselect a task if additional experience in that task is required. 

In Figure \ref{pacmanAgent}(c), we reproduce the continuous state representation CMDP learning curves using value function transfer from Figure \ref{pacmanAgent}(a) and reward shaping from Figure \ref{pacmanAgent}(b). These are denoted in the figure by ``(return-based)", and train on sources until 35\% of the max return is achieved. We compare them against an approach that is identical to ``(return-based)" approaches, but that trains for 5 episodes on a task at a time. These CMDP learning curves are denoted with ``(small fixed)." The results show that agents do not need to train for a long time or to convergence on source tasks, and that our approach can adapt to this hyperparameter setting.





\section{Related Work}
\label{sec:related}


The idea of using a curriculum to train reinforcement learning agents has been around for a long time. Curriculum-like strategies have been used accelerate training in domains from robotics \cite{Asada96, MacAlpine18} to complex multiagent games \cite{Narvekar16, Wu16}. However, typically the curriculum was generated manually, by either a domain expert or naive users \cite{Peng16}.

\emph{Automatically} sequencing tasks into a curriculum is an open problem that has only recently begun to receive attention. Due to the complex nature of the problem, most existing work makes some kind of simplifying assumption. For example, some methods restrict the type of source tasks available for training. For instance, some methods \cite{Florensa17, Held18} only change the initial and terminal state distributions of the final task to create source tasks, while others change the reward function \cite{Riedmiller18, Sukhbaatar17}. In contrast, our method allows source tasks to vary in any way from the target task MDP. However, unlike some of these methods, the set of sources must be provided in advance.

Another class of sequencing methods assume additional domain information about the sources is available to aid in sequencing the tasks. One common assumption is the availability of task descriptors \cite{Silva18, Narvekar17,Svetlik17}, which describe how tasks in a domain relate to one another, and serve as a proxy for task difficulty. Such domain information is typically combined with heuristics, such as transfer potential \cite{Svetlik17}, to guide the selection and training of tasks. In contrast, our approach relies on direct interaction and experience in source tasks to learn a curriculum, and does not use task descriptors. 

Our approach falls within a class of methods that take an MDP-based approach to curriculum generation. It builds off the work of \citet{Narvekar17}, who formulated the idea of curriculum MDPs. Closely related is the work of \citet{Matiisen17}, who model curriculum generation as a POMDP, using a different reward objective and without assuming access to the learning agent's parameters. However, neither of these works actually attempted to learn a policy on the MDP/POMDP. Instead, they opted to use heuristics to extract a single sequence curriculum, rather than the full curriculum policy.







The problem of curriculum learning has similarities to the problem of \emph{source task selection} in transfer learning. In this problem, the goal is to select the best source task from a prespecified set for a given target task. These approaches typically compute a similarity measure between the MDPs of the source and target task \cite{Ferns12,Ammar14b}, or learn a model of transferability that can be applied to novel source-target task pairs \cite{Sinapov15,Isele16}. However, none of these methods have been successfully applied to select a multi-step sequence of tasks.







Finally, curriculum learning has also been explored in the context of 
supervised learning \cite{Bengio09,Graves17, Fan18}. 
Various related paradigms such as multi-task reinforcement learning 
\cite{Wilson07} and lifelong learning \cite{Ammar14} have also been examined. 
The main difference between curriculum learning and these works is that we have full control over the order in which tasks are 
selected, and the goal is to optimize performance for a specific target task, 
rather than all tasks.

\section{Conclusion and Future Work}
\label{sec:conclusion}

In this paper, we showed that a more general representation of a curriculum than previous work, a \emph{curriculum policy}, can be learned. 
The key challenge of learning a curriculum policy is creating a CMDP state representation that allows efficient learning. We extended the original curriculum MDP definition to handle multiple types of transfer learning algorithms, and described how to construct CMDP representations for both discrete and continuous domains to faciliate such learning. Finally, we demonstrated that curriculum policies can be learned on a gridworld and pacman domain. The results show that our approach is successful at creating curricula that can train agents to perform on a target task as fast or faster than existing methods. Furthermore, our approach is robust to multiple learning agent types, multiple transfer learning algorithms, and different CMDP representations. 

One limitation of our approach is that learning a full curriculum policy can take significantly more experience data than learning the target policy from scratch.  An important direction for future work is investigating the extent to which this cost can be amortized by reusing learned curricula for multiple, similar target tasks.  The contributions of this paper are an essential prerequisite for such an investigation.  Another interesting direction for future work is to examine the extent to which the methods presented here generalize to policy-gradient-based approaches and transfer learning algorithms, in addition to the value-function-based algorithms that were used in all of our experiments.

\section*{Acknowledgements} 

The authors would like to thank Felipe Leno da Silva for helping review this paper. This work has taken place in the Learning Agents Research
Group (LARG) at UT Austin.  LARG research is supported in part by NSF 
(IIS-1637736, IIS-1651089, IIS-1724157), Intel, Raytheon, and Lockheed
Martin.  Peter Stone serves on the Board of Directors of Cogitai, Inc.
The terms of this arrangement have been reviewed and approved by the 
University of Texas at Austin in accordance with its policy on
objectivity in research. 
 

\bibliographystyle{ACM-Reference-Format}  
\bibliography{aamas19}


\newpage
\appendix

\end{document}